\documentclass[preprint]{vgtc}               

\graphicspath{{figures/}{pictures/}{images/}{./}} 

\usepackage{times}                     

\usepackage{tabu}                      
\usepackage{booktabs}                  
\usepackage{lipsum}                    
\usepackage{mwe}                       

\usepackage{amsmath}
\DeclareMathSymbol{*}{\mathbin}{symbols}{"01}
\DeclareMathAlphabet{\altmathcal}{OMS}{cmsy}{m}{n}
\usepackage{mathptmx}                  
\usepackage{algorithm}
\usepackage{algpseudocode}
\usepackage{tablefootnote}

\usepackage{subfiles}

\usepackage{mathptmx}                  

\onlineid{1003}

\vgtccategory{Research}

\vgtcinsertpkg

\title{Out-of-Core Dimensionality Reduction for Large Data via\\Out-of-Sample Extensions}

\author{Luca Reichmann\thanks{e-mail: st169765@stud.uni-stuttgart.de}\\ %
        \scriptsize University of Stuttgart %
\and David H\"agele\thanks{e-mail: david.haegele@visus.uni-stuttgart.de}\\ %
     \scriptsize University of Stuttgart %
\and Daniel Weiskopf\thanks{e-mail: daniel.weiskopf@visus.uni-stuttgart.de}\\ %
     \scriptsize University of Stuttgart}

\teaser{
    \centering
    \includegraphics[width=0.98\linewidth]{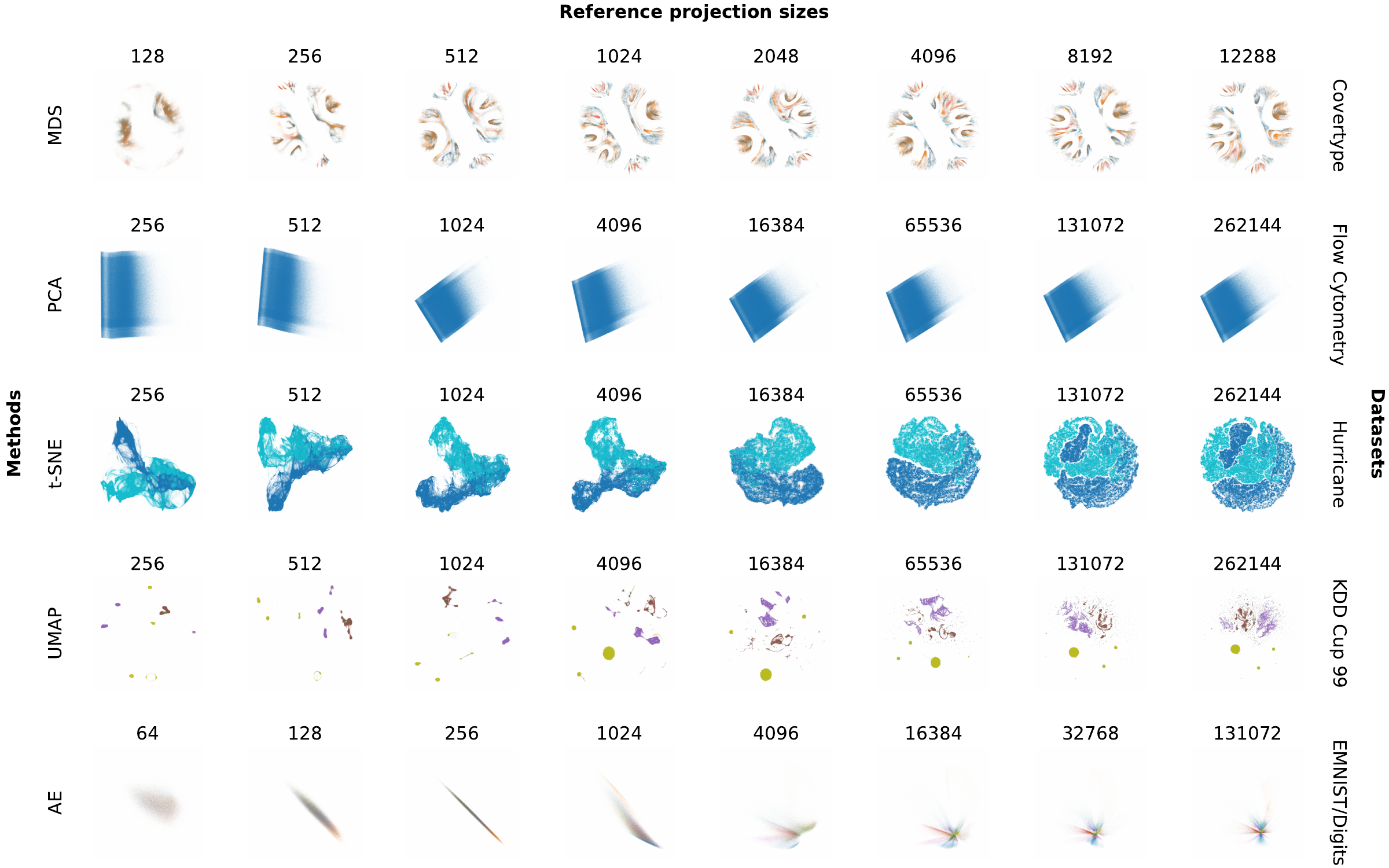}
    \caption{Visualizations of data sets with up to 50 million data points using increasing reference set sizes. We show the results for popular dimensionality reduction techniques: MDS, PCA, t-SNE, UMAP, and autoencoder. The number above each plot shows the reference set size used for creating the initial reference projection.}
    \label{fig:refsize}
}

\abstract{%
Dimensionality reduction (DR) is a well-established approach for the visualization of high-dimensional data sets.
While DR methods are often applied to typical DR benchmark data sets in the literature, they might suffer from high runtime complexity and memory requirements, making them unsuitable for large data visualization especially in environments outside of high-performance computing.
To perform DR on large data sets, we propose the use of out-of-sample extensions. 
Such extensions allow inserting new data into existing projections, which we leverage to iteratively project data into a reference projection that consists only of a small manageable subset.
This process makes it possible to perform DR out-of-core on large data, which would otherwise not be possible due to memory and runtime limitations.
For metric multidimensional scaling (MDS), we contribute an implementation with out-of-sample projection capability since typical software libraries do not support it.
We provide an evaluation of the projection quality of five common DR algorithms (MDS, PCA, t-SNE, UMAP, and autoencoders) using quality metrics from the literature and analyze the trade-off between the size of the reference set and projection quality.
The runtime behavior of the algorithms is also quantified with respect to reference set size, out-of-sample batch size, and dimensionality of the data sets.
Furthermore, we compare the out-of-sample approach to other recently introduced DR methods, such as PaCMAP and TriMAP, which claim to handle larger data sets than traditional approaches.
To showcase the usefulness of DR on this large scale, we contribute a use case where we analyze ensembles of streamlines amounting to one billion projected instances.
}


\begin{document}


\firstsection{Introduction}

\maketitle

Dimensionality reduction (DR) is a popular visualization approach to gain insight into inherent structures of high-dimensional data in different domains. 
However, applying DR to large data sets consisting of millions of instances is challenging or even infeasible for some methods, such as multidimensional scaling (MDS) that scales quadratically in computational effort with the number of data~points.

In this paper, we address the problem of DR for large data,
such as high-dimensional biomedical data consisting of millions of individual observations. 
There are many DR techniques, with principal component analysis (PCA), uniform manifold approximation and projection (UMAP), and t-distributed stochastic neighbor embedding (t-SNE) being popular examples.
Although there are some DR techniques that were developed to overcome the performance issues, such as TriMap \cite{2019TRIMAP} and pairwise controlled manifold approximation projection (PaCMAP) \cite{JMLR:v22:20-1061}, they are still often limited by the available memory.
Existing methods have been evaluated in the literature with data sets consisting of up to 20 million instances~\cite{Belkina2019, tang2016visualizing}, usually taking hours to compute and some requiring hardware exceeding consumer standards, e.g., 128+ GB of RAM.

In this paper, we use similarly sized data sets for comparability but also show that our approach is feasible for data sets comprised of 1 billion instances using around 240\,GiB of memory.  
We resolve the memory problems by using out-of-sample (OOS) extensions of the DR methods.
OOS extensions allow us to successively supplement an existing DR with new samples that are processed with respect to the existing projection.
Depending on the DR technique, the implementation of OOS extensions can vary widely.
While it is trivial to project new data points with PCA, the process is more complex with techniques like UMAP and t-SNE.
Using OOS extensions usually not only lowers memory requirements but also reduces runtime because fewer data point comparisons become possible.
However, this also leads to a loss in DR quality since less data is available to create the initial low-dimensional representation.
Often, this trade-off is primarily influenced by the size of the already existing low-dimensional representation, which is then employed for processing subsequent OOS batches.
We primarily use the term \emph{projection} for the low-dimensional representation in our work, yet other terms, such as \emph{embedding}, are also commonly used in the literature.

We provide a framework description enabling DR to be performed out-of-core by leveraging OOS extensions and making it viable to process data that does not fit into memory.
Our main contribution is an extensive evaluation of OOS extensions for several DR methods using this framework: (metric) MDS \cite{Kruskal:1978:MDS}, PCA, UMAP \cite{mcinnes2018umap}, t-SNE \cite{JMLR:v9:vandermaaten08a}, and autoencoder~\cite{8791727}, spanning DR techniques of different categories used for various purposes. 
The evaluation includes popular quantitative and qualitative measures of the projection quality of the DR methods. \Cref{fig:refsize} shows examples of the DR techniques applied to various data sets.
We also conducted a runtime evaluation of the decrease in computational effort when using OOS extensions compared to the default use of DR techniques.
In addition, we provide a description of a generalized framework for performing DR with OOS extensions to process large data out-of-core.
Our approach can thus be applied to other DR techniques supporting OOS extensions.

Therefore, this paper aims to provide guidance when choosing the number of data points for the initial DR computation to balance the necessary computational effort and the resulting quality when employing OOS extensions with large data sets.

Furthermore, we provide supplemental material~\cite{supplemental} containing additional figures, code to reproduce figures of this paper, and our own implementation of MDS with OOS extension.

\section{Related Work}

We discuss related previous work that addresses the challenges of large data DR in general and specific approaches relying on and evaluating OOS extensions.

\paragraph{DR for Large Data}
There are several efforts to make DR applicable to large data.
While we specifically propose using OOS extensions, others developed DR algorithms with a focus on performance or provided optimized implementations of existing DR algorithms, such as ports to the GPU. 
One of those techniques is Pairwise Controlled Manifold Approximation Projection (PaCMAP) by Wang et al.~\cite{JMLR:v22:20-1061}.
They applied PaCMAP to data sets such as KDD Cup '99, which contains over 4,000,000 data points.
While conventional DR methods such as UMAP or t-SNE either failed because they ran out of memory or took longer than 24 hours to finish, PaCMAP could transform large data sets in a reasonable amount of~time.

Another method is TriMap \cite{2019TRIMAP}, which is also able to process large data sets with slightly higher runtimes.
The authors of the TriMap publication also tested the method with data sets of up to 11 million data points.
In both publications, UMAP and t-SNE tended to fail or exceed the 12-hour time limit at around a data set size of 1 million data points.
While both papers used the UMAP implementation of McInnes et al.~\cite{mcinnes2018umap-software}, TriMap relied on Multicore-TSNE~\cite{Ulyanov2016} and PaCMAP on the Scikit-learn t-SNE implementation~\cite{scikit-learn} for the comparison. 
LargeVis~\cite{tang2016visualizing} was specifically designed to reduce the computational cost of t-SNE.
Similar to the other two methods mentioned, LargeVis was evaluated using data sets with millions of data points.
Zhu et al.~\cite{ZHU202151} improved LargeVis, achieving higher performance with similar visualization quality.

One could also use parametric DR methods, which are intrinsically able to perform OOS projections by learning an explicit function to map points from the high- to low-dimensional space, to perform DR with large quantities of data.
There have been proposals of parametric extensions for multiple non-parametric DR methods, such as t-SNE~\cite{pmlr-v5-maaten09a} and UMAP~\cite{10.1162/neco_a_01434}, which use neural nets to learn the function.
Hinterreiter et al.~\cite{https://doi.org/10.1111/cgf.14834} introduced a framework making parametric extensions generally available for DR techniques, as such extensions are not formulated for every DR method.
While the parametric variants of t-SNE and UMAP provide high-quality projections, it is also noted that parametric versions of DR methods also come with more hyperparameters that have to be adjusted.
In our paper, we use and evaluate OOS extensions instead of parametric extensions of DR methods to project OOS data points.

As implementation details also dramatically influence the feasibility of DR methods, there were different attempts to gain performance enhancements.
A way of achieving this is GPU porting of the techniques, which exist for t-SNE~\cite{8645912, MEYER2022116918} and UMAP~\cite{Nolet_Lafargue_Raff_Nanditale_Oates_Zedlewski_Patterson_2021}.
The authors could achieve a speedup of up to 700 times compared to the single-core Scikit-learn implementation and up to 100 times compared to the original multi-core UMAP implementation.
We also use parallelization for the MDS algorithm by implementing its gradient descent scheme on the GPU.
Typical quality metrics of DR algorithms, such as stress or trustworthiness, often suffer from poor algorithmic scalability \cite{Richer:2022:scalability} as they rely on pairwise comparisons. Therefore, Nolet et al.~\cite{Nolet_Lafargue_Raff_Nanditale_Oates_Zedlewski_Patterson_2021} also ported the trustworthiness computation to the GPU.
The implementation works in batches to save memory.
We continue this approach with the GPU-based implementations of the metrics described in \Cref{sec:methods}.
An alternative approach to dealing with the slow runtime of metrics is to only evaluate a subset of the whole DR computation.
This approach was used by Kobak and Berens~\cite{Kobak2019}.
In contrast, we evaluate the complete DR computation with the GPU-based computation of the metrics to obtain more accurate and reliable insights.

\paragraph{Use and Evaluation of OOS Extensions}
OOS extensions are available for many DR methods that do not intrinsically support the transformation of new data points to the lower dimension with respect to the existing projection.
While it is often mentioned in the literature that the use of OOS extensions might lead to lower quality~\cite{NEVES2022233, 8383983}, a comprehensive study of the effects of using OOS extensions is lacking to the best of our knowledge.

However, there are proposals and corresponding evaluations of individual OOS extensions.
One such extension was presented by Bengio et al.~\cite{NIPS2003_cf059682}, where they introduced an OOS framework by learning the corresponding eigenfunctions.
Similar to our evaluation, they evaluated the framework with different training set sizes.
Gisbrecht et al.~\cite{gisbrecht2012out} proposed kernel-based methods to obtain OOS extensions and demonstrated these for t-SNE.
They presented three different methods and evaluated each of them with different training set sizes.
Yet, their evaluation was limited, as they only tested three different training set sizes.
Zhang et al.~\cite{doi:10.1177/1473871620978209} extended the approach of kernel-based OOS extensions, as they identified weaknesses when processing outliers by introducing so-called bi-kernel t-SNE.
They performed a more extensive evaluation of their approach, as they used multiple metrics to measure the quality of their approach and compare it to other state-of-the-art DR methods.
In addition, they conducted a runtime evaluation.

We base our evaluation on these previous works.
Many different quality metrics are available for quantifying projection quality, focusing on different DR aspects.
Comprehensive studies of such measures are provided by Espadoto et al.~\cite{8851280}, Gracia et al.~\cite{GRACIA20141}, and Nonato and Aupetit \cite{8383983}.
These works comprise over 20 quality metrics categorized into local and global measures.
We utilize a selection of them to quantify the projection quality of our proposed algorithm with metrics from both categories.

\section{Methodology}

Our methodology consists of two parts: a framework with a high-level algorithm to compute DR with OOS extensions for any DR method and our evaluation approach.

\subsection{Computational Framework}
We provide a generalized formulation of the OOS projection framework to perform DR on large data sets of many observations, allowing us to plug in any DR method that has an OOS extension.
The pseudo code is listed under \Cref{alg:subsetprojection}. 

\begin{algorithm}
\caption{Projection with random subset reference}\label{alg:subsetprojection}
\begin{algorithmic}[1]
\State var $n_\text{ref}$ \Comment{reference size}
\State var $n_\text{batch}$ \Comment{batch size}
\Procedure{project}{$X,~\Phi$} \Comment{$X$: data set, $\Phi$: DR method}
    \State $X_{a} \gets$ $n_\text{ref}$ random points of $X$
    \State $Y_{a},~\beta \gets \Phi(X_{a})$ \Comment{$\beta$: learned parameters}
    \State $X_{r} \gets X \setminus X_{a}$ \Comment{$X_{r}$: remaining data}
    \For{$i \in \{1~\dots~\lceil \text{len}(X_r)/n_\text{batch}\rceil\}$}\Comment{project batches}
        \State $X_{b(i)} \gets i^{\,\text{th}}$ subset of $X_{r}$
        \State $Y_{b(i)} \gets \Phi_\beta(X_{b(i)})$ \Comment{parameters $\beta$ stay fixed}
    \EndFor
    \State \textbf{return} $Y_{a}~\cup~Y_{b(1)}~\cup~\dots~\cup~Y_{b(n_\text{batch})}$
\EndProcedure
\end{algorithmic}
\end{algorithm}

A DR method in this framework is a procedure that takes a high-dimensional data set $X$ as input and maps it to a low-dimensional representation $Y$ while learning the set of parameters $\beta$ for the mapping: $Y, \beta \gets \Phi(X)$.
The details and semantics of the parameters depend on the DR method.
For PCA, $\beta$ is the set of eigenvectors; for an autoencoder, it is the set of weights and biases; and for MDS, $\beta$ is the discrete mapping $X \rightarrow Y$ itself.
When the mapping has been learned, the parameters can be reused to map more data points, which we call the OOS projection: $Y' \gets \Phi_\beta(X')$. 
This results in a mapping $X' \rightarrow Y'$ that is consistent with $X \rightarrow Y$.

The parameter $\beta$ determines the amount of information available about the underlying high-dimensional space, providing a way of controlling the trade-off between the projection quality and time necessary to map OOS data.
A set $\beta$ of large cardinality may capture more information but come at higher computational costs for subsequent OOS projections.
Note that the parameters $\beta$ do not change when performing the OOS projection, even though it may be possible in general. Also, the points in the OOS batches cannot ``see'' each other, i.e., relationships between points of the OOS sets are not taken into account.
This is an important detail that provides deterministic behavior independent of batch size and order.

The algorithm starts by creating a small random subset $X_a$ of the data points.
The subset is then used to learn the mapping to low-dimensional space.
The rest of the data $X_r$ is then split up into batches $X_{b(i)}$ and mapped via OOS projection with the learned parameters.
This means that the mapping $X_a \rightarrow Y_a$ is used as a reference for the projection. Hence, we call the initial subset the \emph{reference set} similar to the \emph{training set} in machine learning contexts.
While splitting into batches is unnecessary from a theoretical point of view, practically, this is key to arriving at an out-of-core algorithm.
The batch processing keeps the memory requirements of the individual DR methods low, and it allows for loading data in parts subsequently from external memory.

\subsection{Evaluation}
To evaluate the proposed algorithm, we test it with several DR methods that have an OOS extension.
Our selection of DR methods includes some of the most popular algorithms and covers different data qualities that are preserved or optimized for, as outlined in \Cref{table:drmethods}.
\begin{table}[!h]
\caption{DR methods used in the evaluation of the OOS framework.}
\label{table:drmethods}
\vspace{-3ex}
\begin{center}
\resizebox{\linewidth}{!}{%
\begin{tabular}{llcc} 
\toprule
 Method & Optimizes for & Linear & $\beta$ \\ 
 \midrule 
 PCA & reconstruction err. & yes & eigenvectors \\
 MDS & global distances & no & $X_a \rightarrow Y_a$ \\
 t-SNE & local neighborhood & no & $X_a \rightarrow Y_a$, kNN \\
 UMAP & local nb., global dst. & no & $X_a \rightarrow Y_a$, kNN \\
 Autoencoder & reconstruction err. & no & weights, biases \\
 \bottomrule
\end{tabular}}
\end{center}
\end{table}

In the evaluation, we use a systematic approach to gain insight into the quality, runtime, and memory usage of the OOS framework.
A key factor is the size of the reference set $X_a$ used to learn $\beta$.
Thus, we report results for varying sizes of $X_a$.
We also compare to the baseline without OOS extensions, i.e., where all data points are projected at once.

\subsubsection{Out-of-Sample Extensions}
The selected DR algorithms use different mechanisms to support OOS projection.
PCA and the autoencoder learn a parametric mapping and give an explicit function to map the data, which serves as $\Phi_\beta(\cdot)$.
For metric MDS, we use the process of performing stress minimization for a single point while keeping all others fixed, referred to as \emph{single scaling} in the literature~\cite{wojciech1999incrementalMDS,leon2005programsimilarity}, to perform the OOS projection of a single point.
The gradient for an OOS point $x'$ with respect to the points $x_i$ of the reference set $X_a$ and corresponding low-dimensional points $y_i$ is given as
$\delta = \sum_i \big( 1-d(x',x_i)/\lVert y' - y_i\rVert\big) * (y' - y_i)$, with dissimilarity function $d(\cdot\;,\,\cdot)$.
A similar mechanism has been proposed for t-SNE~\cite{bergman2014tsneOOS}, which leverages informed initialization based on the OOS point's k-nearest neighbors (kNN).
The implementation we employed approximates kNN and uses interpolation-based t-SNE~\cite{Linderman2019fftTsne} to boost performance on large data sets.
To the best of our knowledge, the OOS extension for UMAP is not described explicitly in the literature.
However, similar approximate nearest neighbor mechanics can be found in UMAP's source code~\cite{mcinnes2018umap-software}.

\subsubsection{Data Sets}
We conduct the evaluation with large data sets of up to several millions of data points; see \Cref{table:datasets}.
While these data sets are not particularly large in the sense of large data analysis and the visualization community in general, they are large in terms of what modern DR methods can handle within hours of computation and limited memory.
Although even much larger data sets would be possible with our OOS framework, we chose to stick to these data sets because they have already been covered in the DR literature.
The metric-based evaluation also becomes challenging with respect to runtime with increasing data set size, which is another reason for us to stick to these sizes.
\begin{table}[t]
\vspace{1.45mm}
\caption{Data sets used in the evaluation of the OOS framework.}
\label{table:datasets}
\vspace{-3ex}
\begin{center}
\resizebox{\linewidth}{!}{%
\begin{tabular}{lrcrc} 
\toprule
 Name & Data points & Class count & Dim. & Source \\ 
 \midrule
 EMNIST/Digits & 280,000 & 10 & 784 & \cite{cohen_afshar_tapson_schaik_2017} \\
 Covertype & 581,012 & 7 & 54 & \cite{misc_covertype_31}\\
 Tornado & 2,097,152 & 1 & 3 & \cite{Crawfis03:Tornado} \\
 Flow Cytometry & 3,176,162 & 1 & 23 & \cite{https://doi.org/10.1002/cyto.a.22983} \\
 KDD Cup '99 & 4,898,431 & 23 & 41 & \cite{misc_kdd_cup_1999_data_130} \\
 Higgs & 11,000,000 & 2 & 28 & \cite{Baldi:2014kfa} \\
 Hurricane Isabel & 50,000,000 & 2 & 13 & \raisebox{-1mm}{\tablefootnote{\url{vis.computer.org/vis2004contest/}}$^,$\tablefootnote{\url{vets.ucar.edu/vg/isabeldata/}}} \\
\bottomrule
\end{tabular}}
\end{center}
\vspace{-2ex}
\end{table}%

The evaluation data sets were chosen to cover a wide range of data sources, application areas, and data characteristics. 
The \emph{EMNIST/Digits} data set contains 28$\times$28 pixel images of handwritten character digits split into ten classes. 
\emph{Covertype} shows forest data in the US, where each data point represents a 30\,m$\times$30\,m patch describing the cover type. 
The \emph{Flow Cytometry} data set consists of fluorescence information acquired via the process of flow cytometry. 
\emph{Tornado} is a uniformly sampled version of a synthetic 3D vector field representing a tornado. 
\emph{KDD Cup '99} contains network data observations, including properties such as the protocol type and duration, which are categorized into normal and different types of intrusions.
The data points of the \emph{Higgs} data set represent two classes of signal processes: one that produces Higgs bosons and one that does not (background). 
\emph{Hurricane Isabel} from the IEEE Visualization Contest 2004 consists of time-varying multidimensional data on a 3D uniform grid that represents simulated environmental variables as a hurricane travels across the land.
We only use two time steps out of 48, amounting to 50~million data points.

\subsubsection{Metrics}
\label{sec:methods}
We use common quality metrics to measure the quality of OOS extension techniques.
The metrics are often categorized as local or global ones~\cite{8383983, GRACIA20141}.
As the OOS framework is evaluated with different DR algorithms optimizing for different goals, we use metrics of both categories to provide comprehensive insights.
Moreover, we use qualitative result inspection~\cite{isenberg2013qri} to evaluate the types of quality of the projections that metrics cannot capture.

\paragraph{Global Metrics}

The \emph{stress} metric \cite{kruskal1964nonmetric} is defined as $\sqrt{{\sum_{i \neq j}(d_{ij} - \lVert y_i-y_j\rVert)^2}~\big/~{\sum_{i \neq j}(d_{ij}^2)}}$, where $d_{ij}$ is the measure of dissimilarity of the high-dimensional points $x_i$ and $x_j$ that is compared to the Euclidean distance of corresponding low-dimensional points $y_i$ and $y_j$.
In our case, $d_{ij}=\lVert x_i-x_j\rVert$. Thus, stress measures the preservation of pairwise distances, where small stress indicates good preservation, with 0 being the ideal value.
The metric is not bounded by a maximum value as it is scale-dependent.
The stress calculation suffers from quadratic time and memory complexities due to its dependency on the pairwise distance matrix. Therefore, we perform the calculation in batches on the GPU.

Similar to Geng et al.~\cite{1542257}, the \emph{Pearson correlation coefficient} is used to calculate the correlation between $d_i$ and $\hat{d_i}$, which represents the $i$-th distance entry of the flattened low- and high-dimensional distance vectors.
A higher value indicates a higher correlation, with 1 being the best possible and 0 being the worst possible value.
We chose the Pearson correlation coefficient instead of the Spearman rank correlation coefficient due to performance considerations when evaluating large projections.
As the stress metric is scale-dependent and the Pearson correlation coefficient is not, we can compare how far clusters spread out in the projection with the stress metric.
With the Pearson correlation coefficient, it is then possible to compare the projections independently of how spread out the projection is.  

\paragraph{Local Metrics}
The following local metrics provide a parameter $k$, determining how many neighbors are taken into account when computing the metric value.
As we apply the quality measures to larger data sets than in most other DR publications, we chose to consider larger neighborhoods (i.e., $k=100$) than most other publications and metric implementations.
This results in measuring a neighborhood containing 0.005\% of all points of a data set with 2~million points, separating it from global measures.

The \emph{KNN precision} \cite{Kobak2019} metric measures the accordance between the neighborhoods of a point in the low- and high-dimensional space.
It is defined as $\frac{\sum_i \frac{1}{k} * H_{{k}_i} \cap L_{{k}_i}}{n}$, where $k$ denotes the number of neighbors considered, $n$ the number of data points, $H_{{k}_i}$ the indices of the $k$-nearest neighbors of the high-dimensional point $x_i$, and $L_{{k}_i}$ the indices for the corresponding low-dimensional point $y_i$.
The metric is again bounded to $[0, 1]$, where higher values indicate an increased neighborhood overlap between the low- and high-dimensional representation.

The \emph{trustworthiness} metric \cite{venna2006visualizing} is defined as $1 - \frac{2}{nk(2n-3k-1)} \sum_{i=1}^{n} \sum_{j \in U_k(i)}^{} (r(i, j) - k)$.
The set $U_k(i)$ contains the low-dimensional elements that are in the set of $k$ closest neighbors of $y_i$ but not in the set of the closest neighbors of the same point in the higher-dimensional space.
$r(i, j)$ yields the index of the low-dimensional point $y_j$ in the nearest-neighbor ordering of the neighbors of the point $y_i$.
As a result, the metric measures how well the local neighborhoods are preserved, where higher values indicate better local neighborhood preservation.
Similar to the stress metric, the trustworthiness metric is calculated in batches on the GPU using the cuML library \cite{raschka2020machine}. 

\section{Computational Complexities}
The DR techniques evaluated have different runtime and memory complexities, which could possibly be improved using the OOS extension framework described above. In this section, we discuss asymptotic complexity as a measure of the algorithmic scalability of the visualization techniques~\cite{Richer:2022:scalability}.
Regarding scalability with the number of data points $n$,
MDS is in $\altmathcal{O}(n^2)$ for both time and memory, due to its dependency on pairwise distances \cite{gisbrecht2015}.
Similar to MDS, a time and memory complexity of $\altmathcal{O}(n^2)$ is reported for t-SNE \cite{JMLR:v9:vandermaaten08a}.
According to McInnes et al.~\cite{mcinnes2018umap}, UMAP has an approximate runtime complexity of $\altmathcal{O}(n^{1.14})$.
With respect to $n$, PCA scales linearly for the computation of the covariance matrix (when dominated by $n$), and the eigendecomposition is independent of $n$.
Its memory complexity is only dependent on the number of dimensions.
The time for training an autoencoder scales linearly with $n$~\cite{van2009dimensionality}. 
The memory complexity depends on the encoder's architecture, which is also independent of the number of data points.

Our framework is designed to generically bring the computational effort down, making it possible to run demanding DR methods and large data sets on consumer-grade computers.
The algorithmic time complexity of our framework can be described as 
\begin{equation}
\altmathcal{O}(\Phi^\text{RT}(n_\text{ref})) + \altmathcal{O}(\Phi^\text{RT}_\beta(n_\text{batch}) * \text{batch\_count}),
\end{equation}
where $\Phi^\text{RT}$ is the function that calculates the DR method's runtime and $\Phi^\text{RT}_\beta$ represents the runtime of the method's OOS projection operation.
This is because the reference projection has to be created first, and then each batch is projected sequentially using the information from the reference projection.
The time complexities are dependent on the specific DR method used, yet the time complexity of each batch projection is equal due to constant batch sizes.
This also applies to the memory requirements, as we do not need to compute on the whole data set at the same time.
Consequently, the memory requirements are constantly $\altmathcal{O}(\Phi^{M}(n_\text{ref})) + \altmathcal{O}(\Phi^M_\beta(n_\text{batch}))$ for each individual batch projection, where $\Phi^M$ and $\Phi^M_\beta$ are the functions to calculate the memory requirement of the respective DR method.
Please note that complexities for $\Phi$ and $\Phi_\beta$ are possibly unequal.
For PCA and the autoencoder, the OOS projection via $\Phi_\beta$ is approximately in $\altmathcal{O}(n_\text{batch})$, since the data only needs to be transformed via their explicit projection function.
In MDS, the points in the OOS set are only compared to the reference set but not among themselves, resulting in a time complexity of $\altmathcal{O}(n_\text{ref}*n_\text{batch})$.
Similarly, the t-SNE and UMAP implementations ignore relations within the OOS set.

\section{Experiments}
In the following, we demonstrate how the framework performs with the described data sets and various DR techniques.
We highlight both qualitative and quantitative results by inspecting the projections and including the results of the metric values.
Furthermore, we show how the runtime changes with increasing reference set sizes and how the techniques compare to each other.

\subsection{Setup and Implementation}
Our benchmarks were run on a workstation equipped with an AMD Ryzen Threadripper PRO 3995WX with 64 CPU cores, an NVIDIA RTX A6000 GPU with 48\,GiB of VRAM, and 251\,GiB of RAM.
This is not a consumer-grade computer, and the OOS framework requires much less computing resources. However, a potent machine was required for large projections with regular DR methods and the computation of quality metrics.
The software environment used for running and evaluating the DR techniques was implemented in Python with Numba~\cite{10.1145/2833157.2833162} compiled functions for performance-critical sections.
We used the UMAP implementation of the umap-learn library \cite{mcinnes2018umap-software}, the openTSNE library \cite{Polivcar731877} for t-SNE, scikit-learn \cite{scikit-learn} for PCA, and keras \cite{chollet2015keras} for the autoencoder.

Since an MDS implementation with an OOS extension is missing in popular DR libraries, we provide our own implementation that can be applied in an out-of-core fashion with GPU-accelerated gradient descent, which can be found in the supplemental material~\cite{supplemental}.
We used standard hyperparameter settings if provided by the corresponding DR libraries.
We set the number of iterations to 350 for UMAP, which is the maximum it defaults to for large data sets.
For MDS, we chose a fixed number of 500 iterations with a step size of 
$10^{-4}$
to strike a balance between good quality results and reasonable runtime.
The autoencoder models used in the evaluation were inspired by the models from the literature~\cite{8791727}, where ReLU activation functions are used in the encoder- and decoder layers, linear and a sigmoid activation function in the last layers.

\subsection{Reference Set Size Quality Trade-off}
\begin{figure}[t]
    \centering
    \includegraphics[width=1\linewidth]{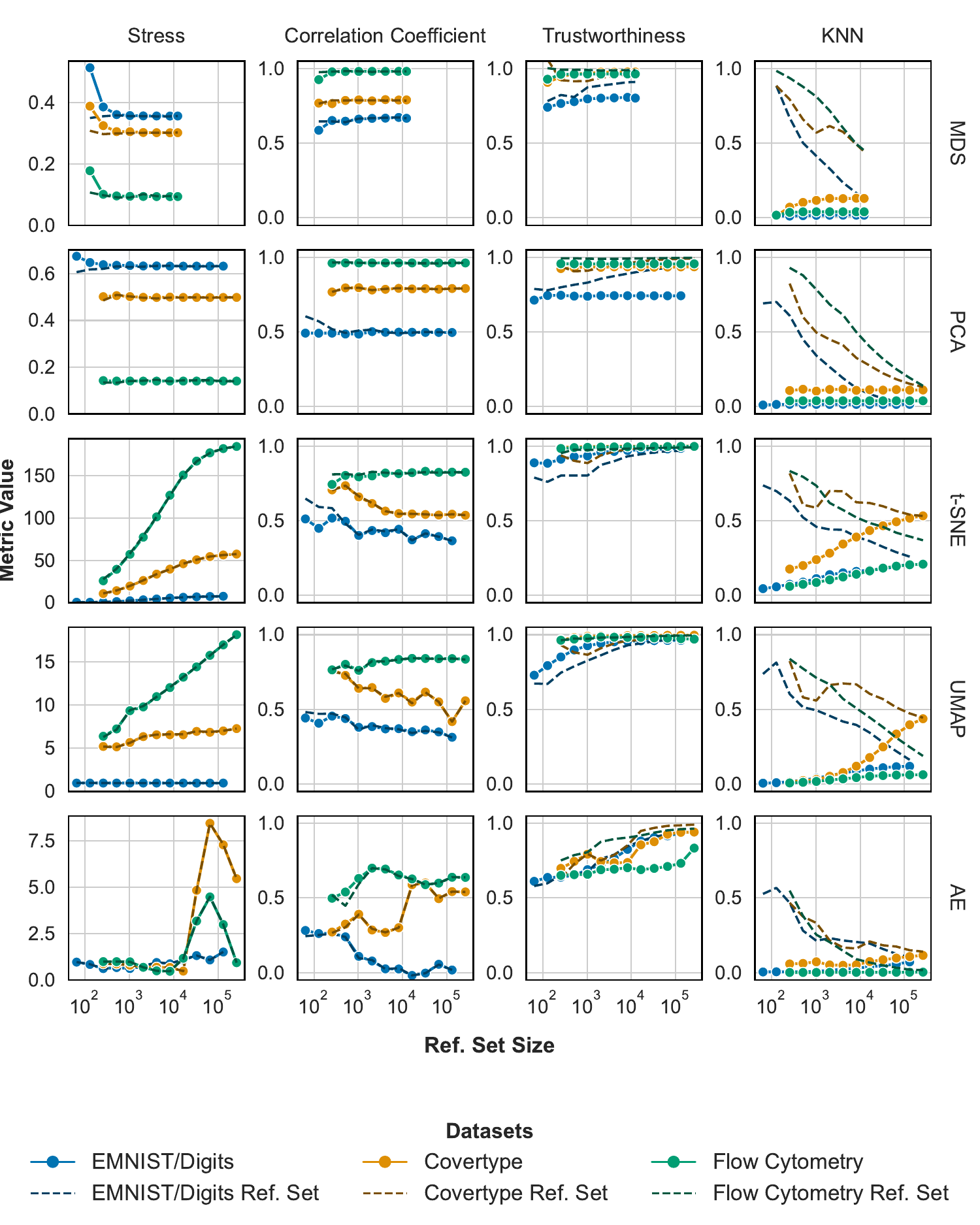}
    \caption{Metric values for increasing reference set size for the EMNIST, Covertype, and Flow Cytometry data sets. The x-axis uses a logarithmic scale for the reference set sizes, and the y-axis displays the corresponding metric values.}
    \label{fig:refsize-metrics}
\end{figure}

In this subsection, we evaluate the projection quality with respect to different reference set sizes.
\Cref{fig:refsize} shows the impact of reference set size for combinations of different DR methods and data sets.
The number in each cell of the figure refers to the reference set size used to create the corresponding projection.
To restrict the runtime of MDS to a reasonable range, we decided to limit the set size to $2^{13}$+\,$2^{12}$ data points.
To substantiate the qualitative observations, we include metric values in \Cref{fig:refsize-metrics}.
Since the metric computations become increasingly slow with large data sets, we present the metric values for the smaller data sets of our collection~(\Cref{table:datasets}): EMNIST, Covertype, and Flow Cytometry.
We include metric values for the reference and whole projections, providing insights into OOS projection performance.

In \Cref{fig:refsize}, it can be seen that the projections, in general, become more consistent with increasing reference set size.
MDS appears consistent pretty early on.
Even with 128 data points, the global overall point arrangement is already visible.
This is also observable in the stress metric results, where the values stabilize at a size of 1,024.
Both the stress and correlation coefficient metrics show a larger discrepancy between the projection of the reference set and the whole projection when using small reference set sizes.
The local measures indicate that the projections also tend to have higher accordance between the low- and high-dimensional neighborhoods with larger reference set sizes. However, local measures are less meaningful for MDS than global measures.

For PCA, the structure of the Flow Cytometry data set projection is already well visible for the smallest reference set.
However, different rotations of the data set are noticeable, showing that the principal vectors still vary for sizes up to 16,384.
The smaller changes in the projections of PCA, when compared to the other methods, are also reflected in the corresponding metric values, where almost no changes are visible.
Nonetheless, with very small reference sets, there are some minor weaknesses, as can be seen in the stress and trustworthiness values of the EMNIST data set (\Cref{fig:refsize-metrics}).

\begin{figure}[t]
\vspace{1.45mm}
    \centering
    \includegraphics[width=1\linewidth]{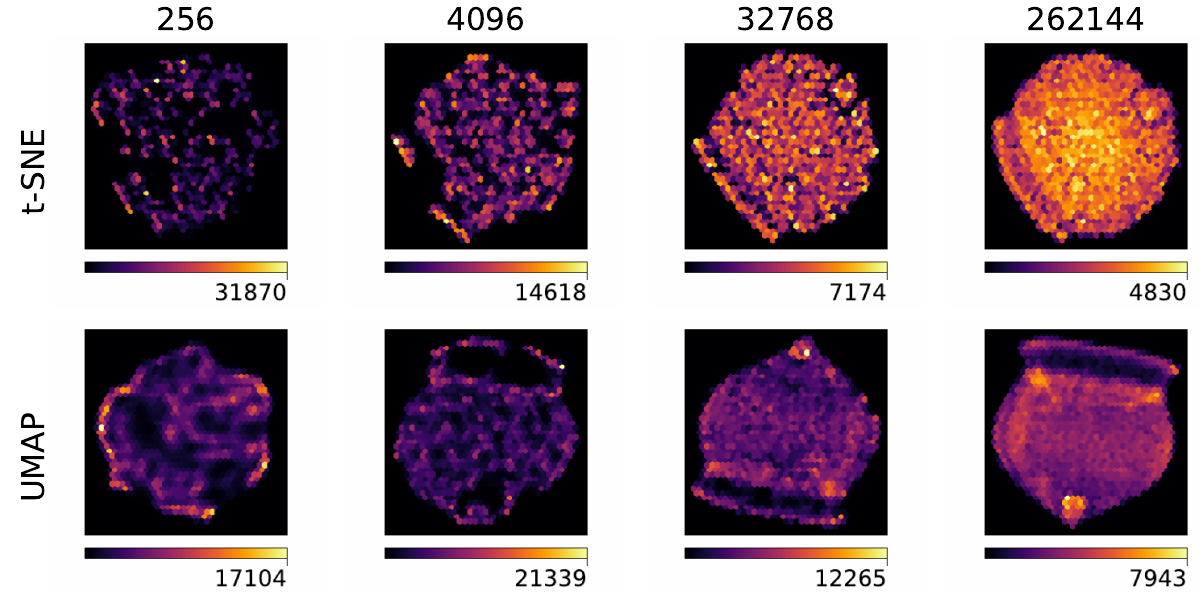}
    \caption{Heat maps of the projections of the Flow Cytometry data set with t-SNE and UMAP. The color bar label shows the number of data points in the most populated area. 
    }
    \label{fig:refsize-heatmap}
\end{figure}

The projections created with t-SNE and UMAP show the most variation throughout the different reference set sizes,
which we think is an effect of overfitting on small reference sets.
Due to their objective of preserving local neighborhoods, they typically also show higher values of trustworthiness and KNN accuracy with small reference sets compared to the other methods (see \Cref{fig:refsize-metrics}).
The KNN metric increases more drastically with larger reference set sizes, which is especially noticeable with the Covertype data set.
The plot also shows that the KNN metric for the reference set and the whole projection converge, which is expected since the reference set size approaches the actual data set size. 
The correlation coefficient measure shows larger differences between the reference set sizes, getting worse for the EMNIST and Covertype data sets with both techniques.
These more drastic changes are also reflected in the projections itself, which continuously change with increasing reference set size.
Both techniques tend to form more distinct, compressed clusters with smaller reference sets.
This can be seen in the purple clusters of the KDD Cup '99 data set, both clusters of the Hurricane data set, and in \Cref{fig:refsize-heatmap}, where we include heat maps of the Flow Cytometry data set created with UMAP and t-SNE.
It can be seen that with both methods, the point distribution in the projections becomes increasingly spread out.
This is also apparent in the heat maps, where the densest regions drop from over 30,000 samples per tile to under 5,000 and from over 17,000 to under 8,000, respectively.
The heat maps also indicate that t-SNE seems to generate more compressed clusters than UMAP, which is likely the consequence of UMAP optimizing more for global relationships than t-SNE.
While it may seem that larger reference sets lead to better global distance relationship preservation with both techniques, the global stress and correlation coefficient measures contradict this observation.

The results of the autoencoder are more inconsistent than those of the other techniques.
With the smallest reference set size, the data points in the projection of the EMNIST data set are arranged in a point cloud, then become line-shaped, and starting from 4,096 data points, the expected distinct separation of classes becomes visible.
This behavior is also reflected in the metric values of the other data sets.
While the other techniques tend to show a slow and steady improvement or decline, there are larger fluctuations in the autoencoder metrics.
This leads to the conclusion that the training process of the autoencoder is influenced more significantly by the size of the reference set than those of the other DR methods.
While the global quality metrics might imply that the projections become worse with larger reference sets, both local metrics show better results.

\subsection{Trade-off Between Reference Set Size and Runtime}
\begin{figure}
\vspace{1.45mm}
    \centering
    \includegraphics[width=1\linewidth]{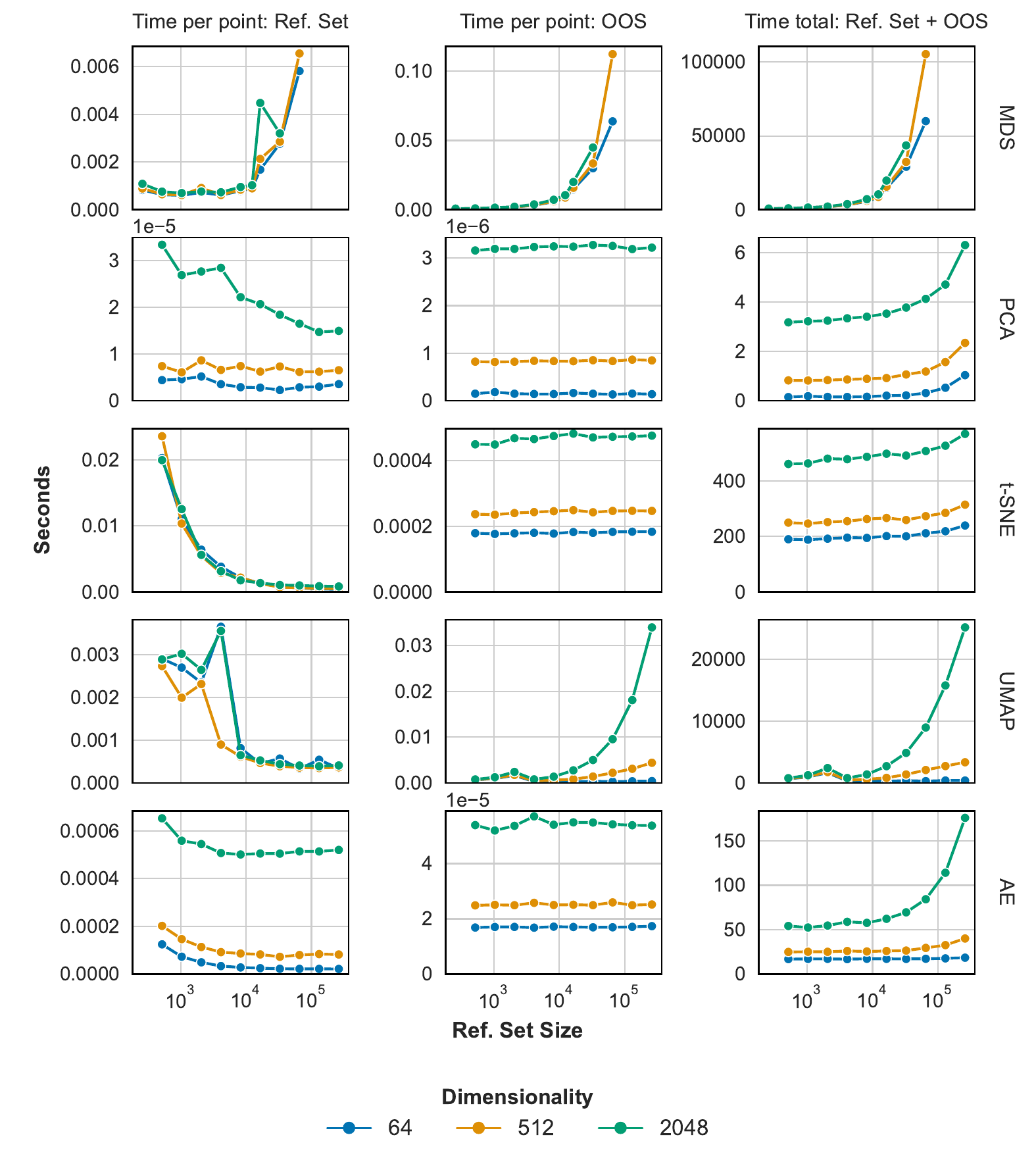}
    \caption{Runtime (in seconds) needed for the projections for varying reference set sizes.
    The x-axis uses a logarithmic scale for the reference set sizes.
    The line colors refer to the dimensionality of the projected data set.}
    \label{fig:refsize-speed}
\end{figure}

\begin{figure}[t]
\vspace{0.5mm}
    \centering
    \includegraphics[width=1\linewidth]{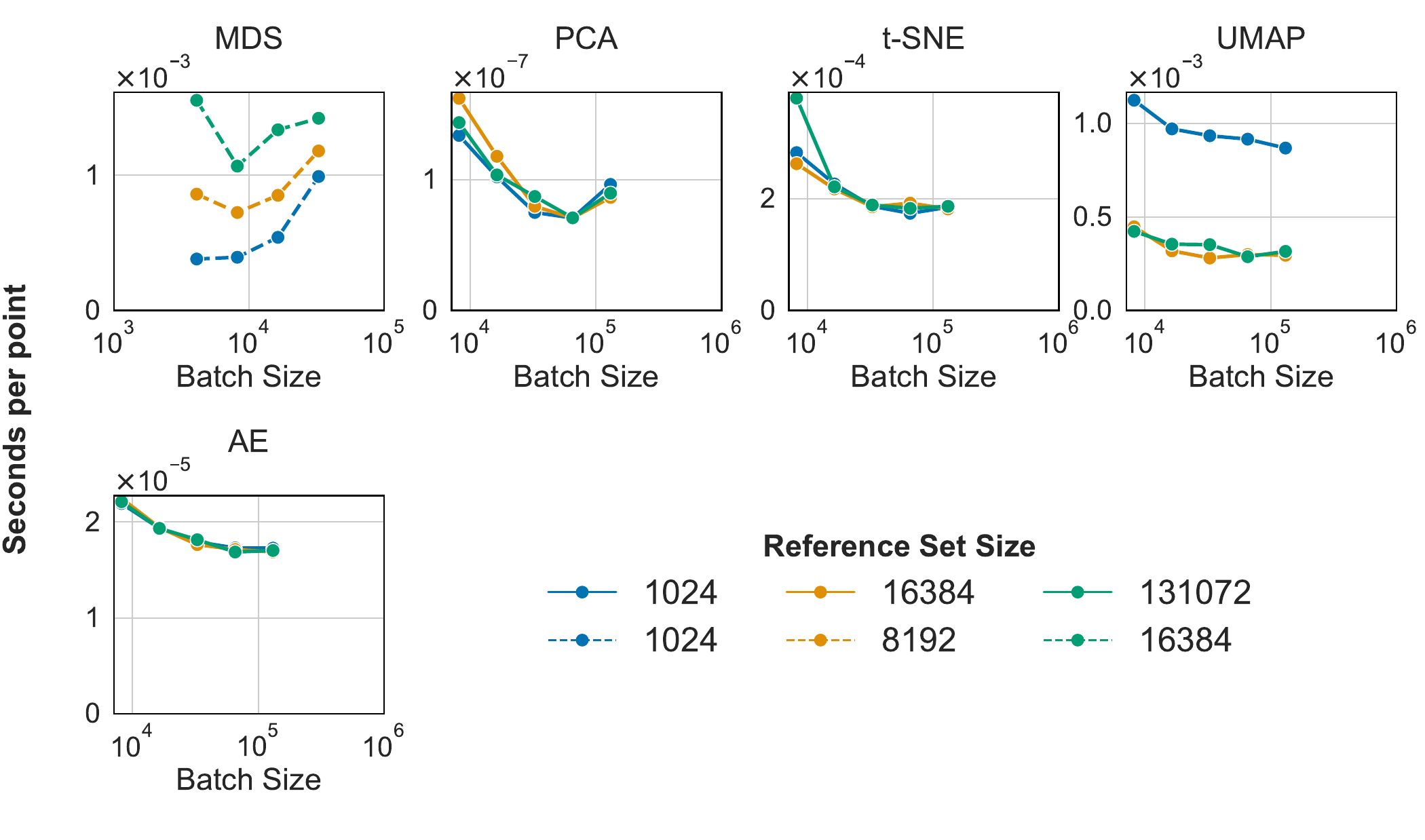}
    \caption{The time necessary to project a point with batches of varying size. The time is measured with the three different reference set sizes per technique (see the legend).}
    \label{fig:batches-speed}
\end{figure}

We measure the time consumed by the DR methods using different reference set sizes.
We use a synthetic data set consisting of four isotropic Gaussian blobs with 1,000,000 instances in total, which is varied in the number of dimensions to capture the relation of dimensionality to runtime. 
In our experiment, we use dimensionalities of 64, 512, and 2,048, which are chosen based on the dimensionality ranges of typical data sets used for DR.
We use reference set sizes similar to those in the quality trade-off experiments.
The OOS projections were performed in batches of 100,000 data points, except for MDS, where we used 1,000 points per batch due to GPU memory limitations.
We include the results of the runtime evaluation in \Cref{fig:refsize-speed} and \Cref{fig:batches-speed}.
The left column of \Cref{fig:refsize-speed} shows the time necessary to compute the reference projection, and the middle column the time needed for the projection of the OOS batches. 
The right column shows the total time necessary for the whole projection. 
We divide the times by the number of points projected in the left and middle columns to get a comparable unit of time per point.

In the runtime graph of MDS, it can be seen that the time to project subsequent batches increases up to over 0.1 seconds per point.
The runtime of the reference set projection shows a similar effect, where the time per point rises from under 2 milliseconds to over 6.
The projection runtime increases with dimensionality.
Yet, the influence is negligible, as the dimensionality only affects the computation of the high-dimensional distance matrix, which is pre-computed once before the optimization process of each OOS~batch.

The per-point projection runtime of OOS batches with PCA does not significantly increase with larger reference sets, which results from only needing to apply a single matrix multiplication on the OOS batches.
We can see that the total runtime slightly increases to over 6 seconds with the largest reference set and the 2048-dimensional data set, which can be attributed to the overhead of processing a larger reference set.

The openTSNE library varies hyperparameter settings depending on the input data, influencing the runtime behavior of the algorithm.
Thus, we decided to always use the approximate neighbor search and employ interpolation-based t-SNE in this experiment to obtain more consistent results.
The runtime behavior shows that with increasing reference set sizes, the mean time to project an individual point steadily and quickly decreases.  
The OOS batch projection runtime is not influenced by the size of the reference set, as the runtimes per point are almost constant with all data set dimensionalities.
Despite that, the overall runtimes increase with larger reference sets.

The runtime behavior of UMAP in the reference set projection process is hardly influenced by the dimensionality.
In contrast, the times necessary for the OOS projections are affected more substantially by the dimensionality of the synthetic data set.
The per-point runtimes for the 64-dimensional data set are low for all reference set sizes, while they increase to over 0.03 seconds for the 2048-dimensional data set with the largest reference set.
The total runtime of the projection is, therefore, also sensitive to the size of the reference set, showing an increase from under 1 second to over 20,000 seconds for the 2048-dimensional data set.

The autoencoder runtimes of the OOS batch projection are comparable to those of PCA and t-SNE.
While the autoencoder itself is slower than PCA and faster than t-SNE, the overall runtime progression is similar, as the times do not increase with larger reference sets.
This is expected since transforming new samples into the low-dimensional space is independent of the size of the reference set once the network is trained. 
Yet, the training process runtime is dependent on the size and dimensionality of the reference.
We can see that the training runtime per point decreases marginally in the first five projections before stabilizing.

In \Cref{fig:batches-speed}, we show the time necessary to perform DR with an individual batch of varying sizes.
We use the same synthetic data set with 32 dimensions.
It can be seen that the time to project one point decreases with the size of the batch with each technique, except for PCA in the last step and MDS with increasing times for larger batches.
The runtime results of MDS and UMAP indicate that they are more sensitive to the reference set size when it comes to the projection of subsequent OOS batches.
This is in accordance with the results shown in \Cref{fig:refsize-speed}, where the per-point times of the projection are almost constant with PCA, t-SNE, and the autoencoder with increasing reference set sizes.
While the batch projection times increase with larger reference set sizes using MDS, UMAP generally took the longest with the small reference sets.

\section{Comparison to Large-scale DR Methods}
\begin{figure}
    \centering
    \includegraphics[width=1\linewidth]{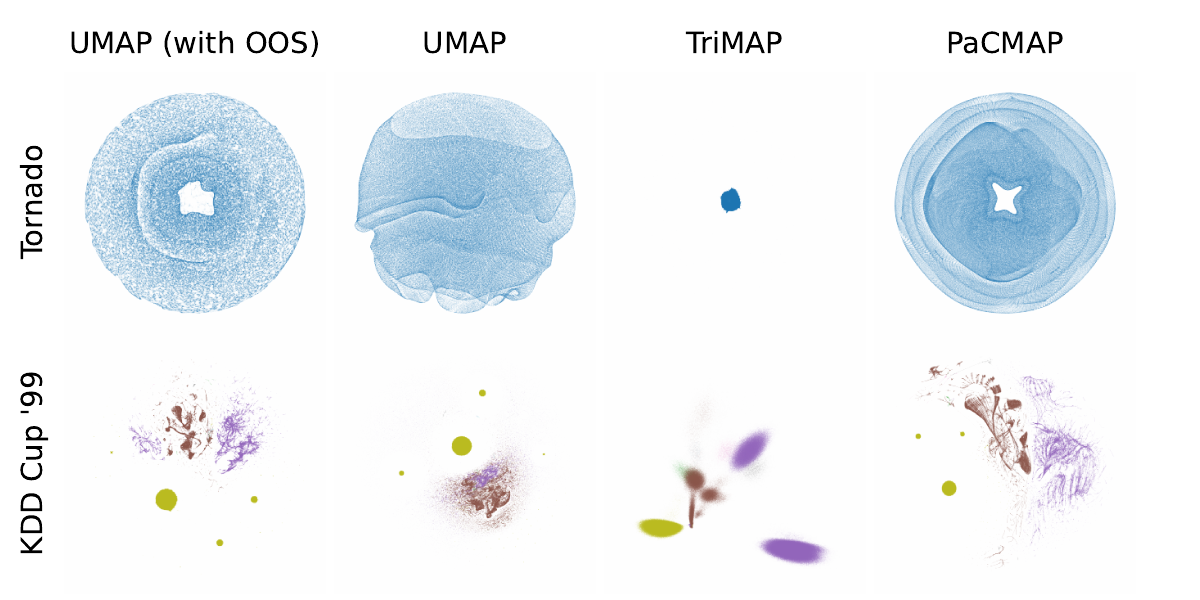}
    \caption{Projections of the Tornado and KDD Cup '99 data sets in the first and second row, respectively. The projections in the first column were created with UMAP and a reference set size of 262,144.}
    \label{fig:state-art-comparison-fig}
\end{figure}

In this section, we compare the OOS framework with UMAP to the large-scale DR methods TriMAP \cite{2019TRIMAP} and PaCMAP \cite{JMLR:v22:20-1061}, both with respect to quality measures and runtime, using the KDD Cup '99 and Tornado data sets.
\begin{table}[t]
\vspace{1.45mm}
    \caption{Results of comparing UMAP and the OOS framework with other state-of-the-art DR methods for large data with the KNN (local) and correlation coefficient (global) quality measures. The best two results per data set and metric are marked in bold. Additionally, the results of running UMAP with 1,000 iterations are included in parentheses.}
    \label{table:state-art-comparison-metric-tab}
        \vspace{-3ex}
    \begin{center}
        \resizebox{\linewidth}{!}{%
        \begin{tabular}{lcccc} 
         \toprule
         ~ & \multicolumn{2}{c}{Tornado} & \multicolumn{2}{c}{KDD Cup '99} \\
         \cmidrule{2-3} \cmidrule{4-5} 
         Method (Ref. Set) & KNN & Corr. Coeff. & KNN & Corr. Coeff. \\ 
         \midrule
         UMAP (4,096) & 0.036 (\textit{0.27}) & 0.165 (\textit{0.14}) & 0.017 (\textit{0.02}) & 0.508  (\textit{0.54}) \\
         UMAP (65,536) & 0.275 (\textit{0.48}) & 0.097 (\textit{0.13}) & 0.094 (\textit{0.11}) & 0.579  (\textit{0.40}) \\
         UMAP (262,144) & 0.363 (\textit{0.51}) & 0.074 (\textit{0.11})& 0.146  (\textit{0.16}) & 0.523  (\textit{0.17}) \\
         UMAP & \textbf{0.370} (\textit{0.35}) & \textbf{0.386} (\textit{0.007}) & \textbf{0.153} (\textit{0.18}) & 0.374  (\textit{0.35}) \\
         TriMAP & 0.271 & 0.098 & 0.001 & \textbf{0.696} \\
         PaCMAP & \textbf{0.522} & \textbf{0.879} & \textbf{0.207} & \textbf{0.768} \\
         \bottomrule
    \end{tabular}}
    \end{center}
\end{table}
In \Cref{table:state-art-comparison-metric-tab}, we include the metric results of the comparison, in \Cref{table:state-art-comparison-time-tab} the corresponding runtimes, and in \Cref{fig:state-art-comparison-fig} the projections of the Tornado and KDD Cup '99 data sets.
The figure shows that the projections of UMAP and PaCMAP are more similar to each other than the projections created with TriMAP, especially visible in the projection of the Tornado data set.
Wang et al. \cite{JMLR:v22:20-1061} also noticed a similar effect with the KDD Cup '99 data set, stating that this is a result of TriMAP primarily preserving global structure.
We can also see a difference in the projection of the Tornado data set between the UMAP projection with OOS projection and pure UMAP, where the latter is missing a hole in the center.
A similar result is obtained for the KDD Cup '99 data set, where the purple points in the projection without an OOS set are arranged around the three yellow dots.
There still seem to be changes with UMAP when using larger reference sets than 262,144 with the given data sets.
It is also conceivable that the global structure of the projections of UMAP with OOS is more similar to the projections of PaCMAP.
The metric values show that PaCMAP tends to perform the best both with regard to global and local quality measures.
UMAP, without using the OOS framework, has the second-best values for the local KNN measure with 0.37 and 0.15.
It can also be seen that in all cases, the KNN metric values increase when the reference set size is increased, where the best values are scored when not using the OOS framework.
However, the correlation coefficient metric does not systematically increase when using larger reference sets, as already observed in previous sections.

\begin{table}
    \caption{Results of comparing the runtimes of UMAP and the OOS framework with other state-of-the-art DR methods for large data, which includes both the projection of the reference set and the OOS batches, if the OOS framework is used.}
    \label{table:state-art-comparison-time-tab}
        \vspace{-3ex}
    \begin{center}
        \resizebox{\linewidth}{!}{%
        \begin{tabular}{l@{\hskip 0.55in}rrr} 
         \toprule
         ~ & \multicolumn{3}{c}{Runtime [in sec.]}\\
         \cmidrule{2-4}
         Method (Ref. Set) & Tornado & KDD Cup '99 & Higgs~~ \\
         \midrule
         UMAP (4,096) & 508.82 & 1,473.15~~  & 3,026.46  \\
         UMAP (65,536) & 514.76 & 1,829.00~~ & 2,693.36 \\
         UMAP (262,144) & 732.46 & 3,591.94~~ & 2,925.83 \\
         UMAP (none) & 1,344.07 & 13,046.95~~ & 5,194.33 \\
         TriMAP (none) & 13,037.25 & 33,155.39~~ & 76,889.32 \\
         PaCMAP (none) & 9,199.30 & 20,661.68~~ & 65,754.48 \\
         \bottomrule
    \end{tabular}}
    \end{center}
\end{table}

TriMAP performs substantially worse with respect to the KNN measure, especially with the KDD Cup '99 data set, where a value of $0.001$ is achieved.
With the Tornado data set, the KNN metric value is higher at $0.271$, yet it is still lower than the values reached with UMAP and PaCMAP.
Amid and Warmuth \cite{2019TRIMAP} also observed lower values of local neighborhood accuracy with TriMAP compared to UMAP and t-SNE with several state-of-the-art DR benchmark data sets.
Similar to their findings, TriMAP also performs well with respect to the global correlation coefficient quality measure in our experiment with the KDD Cup '99 data set, where a value of $0.696$ is achieved compared to $0.768$ with PaCMAP, which is the highest value with this configuration.
However, TriMAP yields lower correlation coefficient values with the Tornado data set, where it performs slightly worse than UMAP and significantly worse than PaCMAP (i.e., 0.098 vs.\ 0.879), which correlates with the unproportional positioning of outlier data points visible in the projection.
The comparison between the three DR methods shows that PaCMAP generally performs the best when compared to TriMAP and UMAP.
TriMAP tends to be inconsistent with its results, while UMAP performs as assumed, where steady improvements to projection quality can be expected.
We can also see that the quality of the projection usually increases for larger reference sets with UMAP, especially with local quality measures.
This shows that with our OOS approach, the reference set size should be chosen as large as possible while considering memory and runtime requirements, as the projection is usually of higher quality.
 
\section{Use Case}
In this section, we apply the OOS framework to an example of a large-scale data set.
Our test data set consists of 1~billion streamlines that we generated on the \emph{fluid simulation ensemble for machine learning}~\cite{Jakob20}.
From the 8,000 flow fields, we used the first 1,000. For each flow field, we generated 1M streamlines using fourth-order Runge-Kutta (RK4) integration with step sizes $\{0.1, 0.2, 0.4, 0.8\}$ and 32 integration steps each.
Our intention was to use DR to get insights into flow behavior and to be able to characterize the flow fields.
To make the streamlines comparable, we translated their origins to zero and rotated them to consistently start in the positive $x$ direction before applying DR.
To build a reference set, 100,000 streamlines were generated at random locations from a small portion of 50 flow fields of the ensemble.
The reference projection was then obtained using t-SNE, where we selected the perplexity parameter so that many clear clusters were forming.
We used the Euclidean distance between the vectors consisting of the streamlines' $x$- and $y$-coordinates $(x_1, y_1,~...~, x_{32}, y_{32})$ as the similarity measure between two streamlines.
Since the 1~billion streamlines that we aimed to OOS project would take approximately 240\,GiB of memory, we chose to generate them on-the-fly and directly pass them in per-flow-field batches to t-SNE for projection.
Thus, we did not need to sacrifice any disk space and could also simulate an in-situ application of the OOS framework.

\begin{figure}[t]
    \centering
    \includegraphics[width=\linewidth]{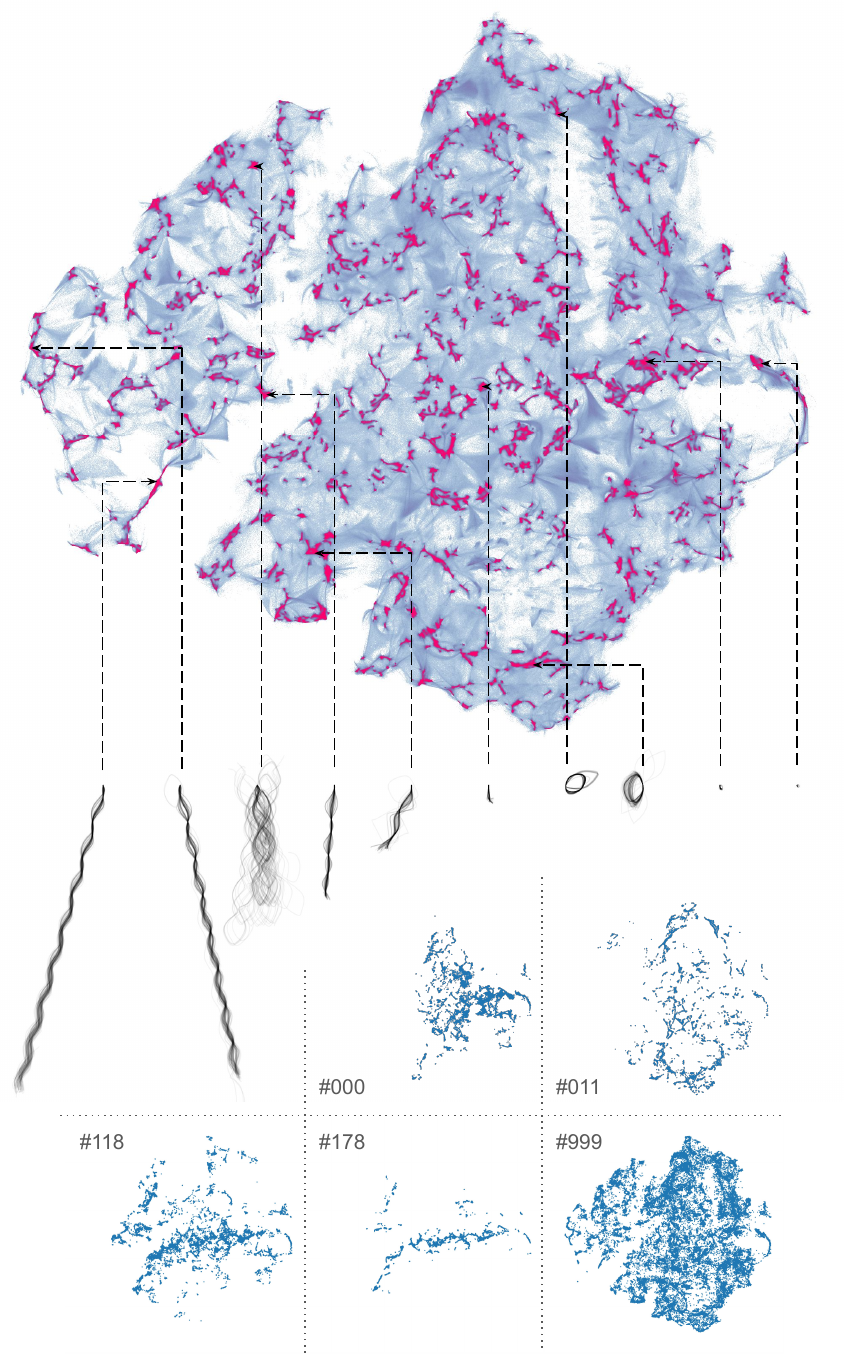}
    \caption{(Top) Heat map of all t-SNE-projected streamlines, where blue indicates low density and red high density. (Middle)
    For different clusters of the projection, 100 of the corresponding streamlines. (Bottom) Scatterplots of subsets of the projection that correspond to different flow fields of the ensemble, showing different coverage of the clusters.}
    \label{fig:casestudy}
\end{figure}

\Cref{fig:casestudy} shows a heat map of all projected streamlines with log-scaled color mapping.
The projection exhibits many little clusters that correspond to similarly shaped streamlines.
A few of them are illustrated below the heat map.
From examining these patterns, we get an idea of the projection's semantics, where very elongated trajectories can be found on the left, and trajectories with little extent where flow is almost stationary are located on the right.
In between, we can find medium-length oscillating streamlines, where clusters differ in orientation and amplitude of the lines, and also elliptical patterns with clockwise and counterclockwise winding.
At the bottom of the figure, there are scatterplots of a selection of flow fields that only contain their respective projected streamlines.
What can be observed is that they differ in coverage of the space, i.e., different flow fields exhibit different kinds of streamlines and allow us to characterize them based on the areas of the projection that they cover.
The projection could be effectively used to explore the flow ensemble when integrated into a visual analytics system, giving an overview from a streamline perspective and serving as a starting point for analysis.

The OOS projection of a batch of 1M streamlines took, on average, 11.8 minutes, of which around 7.4 minutes were used for actual optimization, and the rest of the time was spent on nearest neighbor search and computation of affinities between reference and OOS set.
The whole process of reading each flow field from a file, generating the streamlines, and performing the OOS projection needed 200 hours of computation time, but since the batches are easily parallelizable, we ran four concurrently, resulting in 2 days and 2 hours to obtain the whole projection.
However, upon examining the logs, we observed that improvements in the Kullback-Leibler divergence were stalling early during the optimizations, meaning that the quality gain of using many iterations was negligible, and we could have used only 400 instead of 1000 iterations, which would have saved us 1/3 of the time.

\section{Discussion}
In this section, the results of the evaluation of the OOS approach are discussed, along with limitations and possible future work.

\subsection{Discussion of Results}
We found that the OOS extensions deliver varying results depending on the DR technique.
While techniques like MDS or UMAP are sensitive to the size of the reference set both in runtime and quality, other methods like PCA are not.
Despite these differences, we observed a general increase in quality with respect to metrics measuring local neighborhood preservation.
Depending on the technique, the results of global quality measures were mixed.
While improvements with DR methods optimizing for preserving global relationships can be seen, both the metric values of t-SNE and UMAP dropped for larger reference sets.
This does not necessarily indicate worse quality as the techniques work better toward their goal to visibly separate clusters, as can be seen in the t-SNE projection of the Hurricane data set.
We also noticed that especially the projections of UMAP still change substantially with some data sets when the reference set size is larger than 262,144, which has to be taken into account when using the OOS extension.

We observed improvements in runtime with all techniques when using smaller reference sets.
Again, the results differ depending on which DR algorithm (i.e., what is learned with $\beta$) and data set are used.
Parametric methods, such as the autoencoder or PCA, can typically perform OOS projections quickly, as the learned function can just be applied to the OOS batches again.
Substantial improvements in runtime were achieved when using the OOS approach with the implementations of MDS and UMAP, especially with very high-dimensional data sets.

Another benefit of using OOS extensions is that memory limitations are no longer an issue, as only the reference set and the individual OOS batch have to be stored in memory simultaneously.
While we did not conduct a systematic study on the memory requirements of our approach, we were able to project the KDD Cup '99 data set with 4,898,431 data points with 41 dimensions using our GPU implementation of MDS on an NVIDIA RTX A6000 GPU, which would not easily be possible without an OOS extension due to memory constraints.
Failures with this data set due to memory restrictions with techniques like UMAP and LargeVis were reported by Wang et al. \cite{JMLR:v22:20-1061}, which we also were able to circumvent.
While PaCMAP typically performs the best, UMAP with OOS typically yields better, more reliable results than TriMAP.
Noteworthy is also that the projections of both data sets using the UMAP OOS extension are arguably more similar in global structure to those of PaCMAP than UMAP without OOS extension.

We have also seen that speed is not an issue with UMAP with the hardware used in our state-of-the-art comparison, as the UMAP implementation, in fact, performed better than both TriMAP and PaCMAP.
Yet, the runtime was even further reduced using the OOS approach.
Based on our evaluation and experience from the use case, we recommend that the reference projection should be tested before applying the OOS projection to the whole data set.
By testing, we mean examining the emerging structures and checking if the analysis you aim to do is possible on the small reference subset.
We recommend trying to find out which data characteristics are represented by the different areas of the projection to get an idea of what insights you can expect to find in the overall projection later, so you only spend time and compute resources from promising starting points.

\subsection{Limitations}
The evaluation of the proposed OOS framework is quite time-consuming, mainly due to the computation of metrics but also due to the repeated experiments for different reference set sizes, so we needed to cut corners.
For instance, we did not tune hyperparameters of the DR methods for the specific data sets, e.g., we did not test different perplexities for t-SNE or the number of neighbors for UMAP, but we used default parameters most of the time.
Also, for the autoencoder, we used off-the-shelf architectures suggested in the literature.
To keep runtimes foreseeable, we used a fixed number of iterations and did not employ convergence criteria for terminating optimizations.
Thus, the results we are showing are not benchmark results to compare methods against each other because there was no mechanism in place to achieve the best quality with each DR method. Therefore, the produced projections and timings are not on a competitive level.

Regarding the limitations of the approach, we think that the reference set's projection quality determines the overall quality that is achievable.
We did not evaluate this explicitly, but \Cref{fig:refsize-metrics} supports this hypothesis.
When comparing the metric of the complete projection with the reference projection, it can be seen that they are on par or converging.
In our experiments, the reference sets were selected randomly, so it is possible that certain patterns of the data may not be included in the reference and, therefore, cannot be uncovered for the OOS data.
A more informed selection of the reference set could lead to improved results if the reference, therefore, becomes more representative of the data.
Another limitation by design is that relationships between OOS points are ignored and that the underlying projection model ($\beta$ parameter) is not updated with OOS points.

\subsection{Future Work}
There are several other aspects that we want to examine in detail in the future.
To get a comprehensive evaluation of the OOS framework, we need to expand our selection of data sets to cover an even wider range of domains, sizes, and dimensions.
An extensive benchmark suite for DR methods with OOS extension would be a valuable contribution for the community to get detailed insights into achievable quality and performance for various data sets.
We also seek opportunities to apply the OOS framework as part of a visual analytics system and perform a design study to help solve real-world large data problems. 

\section{Conclusion}
We formulated a generic algorithm for performing out-of-core DR by leveraging OOS extensions and showed that the proposed approach makes it feasible to project large data sets.
The runtime performance and resulting quality of selected DR methods were examined for varying reference set sizes to fathom the trade-off between speed and quality.
We found that some methods were more sensitive to reference set size and needed larger sets to produce consistent projections, while others produced similar projections with small reference sets already.
With a use case, we showcased the feasibility of the OOS framework to produce usable results for a dataset as big as having one billion instances.

\acknowledgments{
This work was funded by the Deutsche Forschungsgemeinschaft (DFG, German Research Foundation)---Project ID 251654672---TRR 161 (Project A01).
}

\bibliographystyle{abbrv-doi-hyperref}
\bibliography{template}

\end{document}